\documentclass[letterpaper, 10.5 pt, conference]{ieeeconf}  

\pdfoutput=1

\IEEEoverridecommandlockouts

\usepackage{graphicx} 
\usepackage{subcaption}
\usepackage{amsmath} 
\usepackage{amssymb}  
\usepackage[noend]{algpseudocode}
\usepackage{algorithm, multicol}
\usepackage{cancel}
\usepackage{siunitx}
\usepackage{xcolor}
\usepackage[numbers]{natbib}
\usepackage{hyperref}
\usepackage{comment}
\usepackage[normalem]{ulem}
\usepackage{booktabs}
\usepackage{multirow}
\usepackage{float}
\usepackage{times}
\usepackage[numbers]{natbib}
\usepackage{multicol}
\usepackage{amssymb}
\usepackage{amsmath}
\usepackage{graphicx}
\usepackage{dblfloatfix}
\usepackage{xcolor}
\usepackage{verbatim}
\usepackage[font=small,skip=1pt]{caption}

\hypersetup{
  pdfinfo={
  pdfproducer={Latex},
  Title={The Past and Present of Imitation Learning},
  Subject={Imitation Learning},
  Author={Nishanth Kumar},
  }
}

\begin{document}
\def\UrlBreaks{\do\/\do-}

\title{\LARGE \bf
The Past and Present of Imitation Learning \\
A Citation Chain Study
}

\author{%
Nishanth Kumar
\\
\small nkumar12@cs.brown.edu
\thanks{Brown University Department of Computer Science}%
\thanks{This writeup is intended as a Final Project submission for CSCI 1951M: The Great Ideas in Computer Science, Fall 2019}
}
\maketitle

\thispagestyle{empty}
\pagestyle{empty}

\IEEEpeerreviewmaketitle

\section{Introduction}
\label{sec:intro}
Imitation Learning is a promising area of active research. Early research in 'programming by example' began in Software Development  \citep{Halbert:1984:PE:911909} before attracting the interest of Robotics and Artificial Intelligence (AI) researchers, who began using the terms 'Learning from Demonstration' and 'Imitation Learning' to describe their line of work. Over the last $30$ years, Imitation Learning has advanced significantly and been used to solve difficult tasks ranging from Autonomous Driving \citep{Pomerleau:1989:AAL:89851.89891} to playing Atari games \citep{Atari}. In the course of this development, different methods for performing Imitation Learning have fallen into and out of favor. In this paper, I will explore the development of these different methods and attempt to examine how the field has progressed.

I will be discussing 4 landmark papers that sequentially cite and inform each other. In discussing these papers, I will focus on their 'big ideas' and how each idea influenced those that came after it. In order of their publication date, these are:
\begin{enumerate}
    \item \textit{'ALVINN: An Autonomous Land Vehicle in a Neural Network'} by Pomerleau \citep{Pomerleau:1989:AAL:89851.89891}
    \item \textit{'Apprenticeship Learning via Inverse Reinforcement Learning'} by Abbeel and Ng \citep{Abbeel:2004:ALV:1015330.1015430}
    \item \textit{'Generative Adversarial Imitation Learning'} by Ho and Ermon \citep{GAIL}
    \item \textit{'A Divergence Minimization Perspective on Imitation Learning Methods'} by Ghasemipour, Zemel and Gu \citep{ghasemipour2019divergence}.
\end{enumerate}

Before discussing the papers themselves, I will provide some essential context by describing the fundamental problem that Imitation Learning attempts to solve.

\section{Background: The Problem of Imitation Learning}
\label{sec:background}
\footnote{Material in this section adapted from \citet{levine}}
At a high level, 'Imitation Learning' attempts to learn a general skill after observing some demonstrations of an expert performing the skill. Crucially, this learned skill is expected to generalize to situations where the learner has not observed the expert's actions.

More formally, let $S_{t}$ represent a set of observations obtained from sensors at time $t$ that constitute the \textit{state}. Let $a_{E_{t}}$ represent a set of actions (eg. voltage sent to motors of a robot) taken by the expert at time $t$. Let $T$ represent a trajectory that is a set of $n$ state, expert action pairs [$(S_{t}, a_{E_{t}})$ for $0 \leq t \leq n$]. Imitation Learning essentially assumes it is given $m$ of these trajectories as input. The expected output is a policy $\pi_{\theta}(a_{t},s_{t})$ (where $\theta$ represents a set of policy-parameters) that maps states to actions. Specifically, it is taken to be a probability distribution that is a function of actions to be taken and states observed at each time step. Crucially, Imitation Learning hopes that the learned policy's distribution of actions will match the experts distribution of actions (i.e, the distribution that $a_{E_{t}}$ are drawn from).

\section{Pomerleau's Autonomous Driving Paper}
\label{sec:autodrive}
Given the substantial research into and success of Machine Learning methods, the most natural thing to do might be to frame the Imitation Learning problem as a 'Supervised Learning' \footnote{For more information on Supervised Learning, refer to Wikipedia's comprehensive and well-written article \citep{SLWiki}} problem. This can be done rather simply: use any Supervised Learning algorithm to learn a function mapping states ($S_t$) to actions ($a_t$) after having trained on the collected trajectories of states observed and expert actions [$(S_{t}, a_{E_{t}})$ for $0 \leq t \leq n$]. This is the essence of Pomerleau's approach to using Imitation Learning for Autonomous Driving.

  \begin{figure}[ht]
      \centering
      \includegraphics[width=0.5\textwidth]{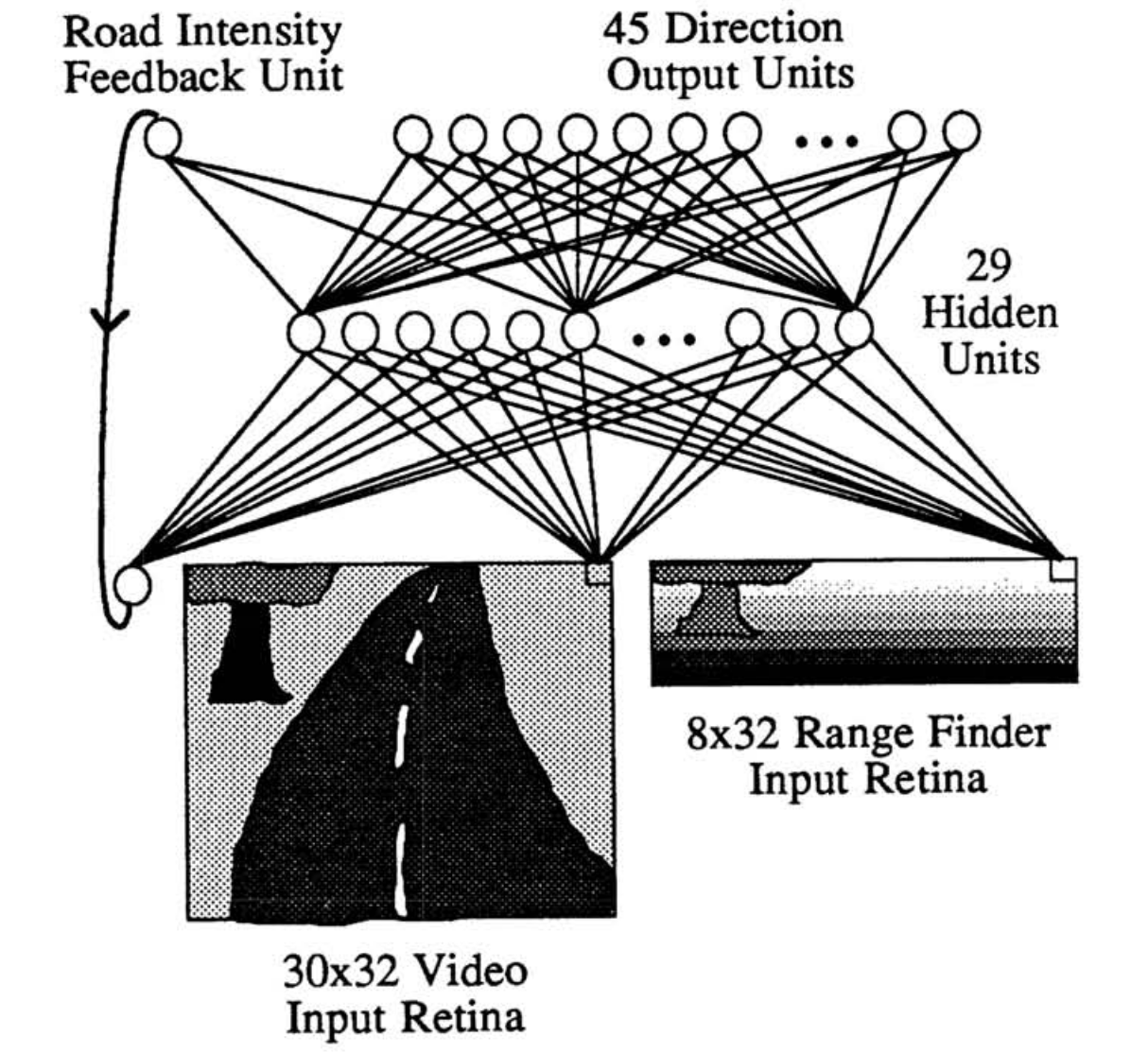}
      \caption{An image from Pomerleau's paper depicting his neural network.}
      \label{fig:PomerleauNetwork}
  \end{figure}

As can be seen from Fig. 1, Pomerleau used a Deep Neural Network (DNN) to perform his Supervised Learning of the expert actions. The network took two inputs: video and data from a laser range finder mounted on the car. It used these to produce a 45-unit long vector representing the curvature that the vehicle would need to travel along to reach the center of the road. It also produced an output indicating how much the road visually contrasted its surroundings. This output was fed back into the network as an input to improve its prediction accuracy.

Unlike many contemporary DNN's, Pomerleau didn't use a real-world dataset to train his DNN because it was infeasible for him to collect, store and process the necessary amount of data. Instead, he created a 'road simulator' that would produce realistic images of roads with added noise and varying lighting conditions. Pomerleau's simulator also generated laser range-finder input corresponding to his images. Finally, the simulator would produce 'expert' actions based on knowing the ground-truth curvature of the road it produced.

Pomerleau's approach exhibited rather impressive performance. His trained DNN was able to drive CMU's NAVLAB (a modified Chevy van equipped with sensors and computers) at a speed of $0.5$ meters per second through a $400$ meter area of CMU's campus under sunny conditions. This performance was comparable to that achieved by the state-of-the-art hand-engineered vision and navigation algorithms of the time. Given that this happened in $1994$ - a time when most computers had a RAM of approximately $4$ MB and DNN's were widely believed to be useless - Pomerleau's results are truly remarkable.

Even though Pomerleau only tested his algorithm on one small environment in optimal weather conditions, his results are still noteworthy. After having developed his road simulator, Pomerleau was able to accomplish in "$30$ minutes of training time" what took CMU's Vision and Navigation groups \textit{months} of hand-tuning their algorithms. This work was one of Imitation Learning's first major successes and it seems to have catalyzed a strong research interest in the field \citep{BALOCH1991271} \citep{BOJARSKI} \citep{Abbeel:2004:ALV:1015330.1015430}.

\section{Abbeel's Inverse Reinforcement Learning Paper}
\label{sec:IRL}
Inspired by the success of \citet{Pomerleau:1989:AAL:89851.89891} and others in using Imitation Learning to solve complex problems, this paper sought to view the Imitation Learning Problem through a different lens. While almost all previous work had framed Imitation Learning as a Supervised Learning problem, this paper leveraged Reinforcement Learning (RL).

Roughly, RL attempts to learn a policy $\pi$ that maps states that an agent could be in ($s$) to actions to be taken from that state ($a$) at every time step. Crucially, the policy is learned such that the actions it takes maximize some reward function $R(s,a)$. At the time that this paper was written, a number of RL algorithms existed to learn this policy \footnote{For a more thorough treatment of RL, refer to the Introduction and initial chapters of \citet{Sutton&Barto}}.

It is important to note that RL algorithms require a reward function ($R(s,a)$) to produce the policy $\pi$. Abbeel and Ng's key idea in this paper is that such an $R$ can be derived from the expert trajectories that Imitation Learning assumes as input. This is what they call 'Inverse Reinforcement Learning' (IRL). Once this reward function ($R_{E}(s,a)$) is obtained, one can use any RL algorithm to obtain a policy $\pi_r$ that maximizes this reward function. If we assume that the expert was attempting to maximize $R_E$, then the policy $\pi_r$ returned by the RL algorithm will (roughly) match the expert's policy. In this way, the Imitation Learning problem can be solved by first using IRL to derive a reward function and then using RL to obtain a policy.

  \begin{figure}[ht]
      \centering
      \includegraphics[width=0.5\textwidth]{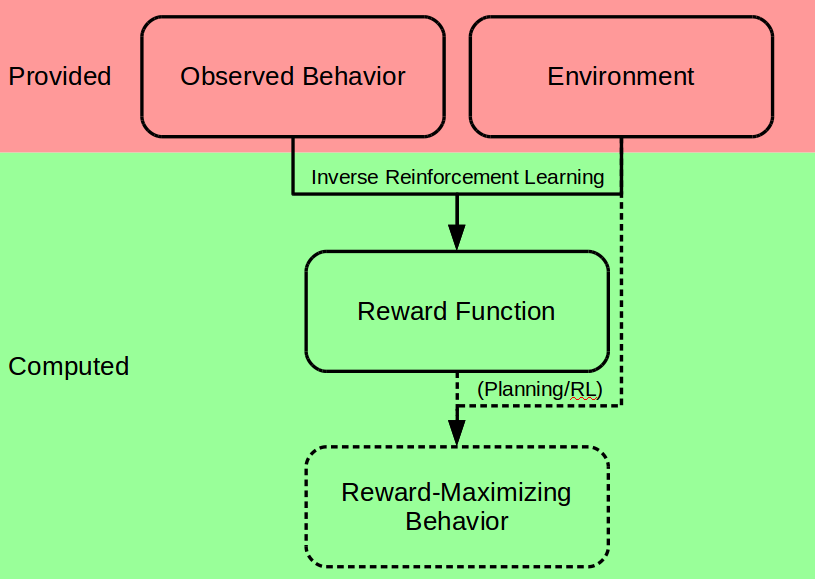}
      \caption{A visual illustration of how Imitation Learning can be performed by the combination of RL and IRL, obtained from \citep{IRLFig}. Since RL uses a reward function as input and outputs optimal behavior, the method that takes an expert's optimal behavior as input and outputs a reward function is termed 'Inverse' RL.}
      \label{fig:overview}
  \end{figure}

Unfortunately, Abbeel's IRL method cannot take as input \textit{only} the expert trajectories. It also requires some set of features $\phi_s$ over states that it can use to learn the reward function. In the case of driving a car, these features might be specific aspects we might want a car to optimize such as whether it has just collided with a car, whether it is in the middle of a lane, etc. Thus, Abbeel's method requires this extra human-specified input in addition to the expert trajectories that \citet{Pomerleau:1989:AAL:89851.89891}'s method operates on directly.

Abbeel is able to use his method to train a car to perform various complicated behaviors in simulation. He demonstrates 5 different learned behaviors that follow policies ranging from simply avoiding all other cars on the road to only driving within the right lane and going off-road to overtake other cars to simply intentionally crashing into the fist car detected.

Overall, this paper introduced a different way of viewing the Imitation Learning problem than that used by \citet{Pomerleau:1989:AAL:89851.89891} and most others before it. It is important to note that this paper did not claim that using IRL and RL to perform Imitation Learning is necessarily \textit{better} than using Supervised Learning, it is simply \textit{different}. One aspect of Abbeel's method that was appealing to many in the community is that it learns an explicit reward function. By inspecting this, one can determine the quantities the agent is attempting to optimize and thus gain some understanding of \textit{why} the policy is taking the actions it takes. In this manner, Abbeel's method is more human-understandable and explainable than the Supervised Learning approaches that came before it.

\section{Ho's GAIL Paper}
\label{sec:GAIL}
In the years after the publication of \citet{Abbeel:2004:ALV:1015330.1015430}'s IRL paper, a significant research interest developed in using IRL methods to learn complex, real-world skills. It was discovered that IRL methods are less error-prone than Supervised Learning methods for Imitation Learning. Since the Supervised Learning methods are trained to mimic decisions the expert made at each time-step, small deviations made early-on in the trajectory can lead to compounding errors later in the trajectory. IRL methods, on the other hand, learn the expert's reward function and are thus able to correct small errors simply by optimizing for the maximum reward. Intuitively, IRL methods have an explicit notion of the expert's goal and can thus optimize for this while direct Supervised Learning are simply attempting to copy what the expert did and generalize slightly \citep{GAIL}.

However, IRL methods were found  to be slow and inefficient with data. This is because they need to iteratively estimate the expert's reward function \textit{and} run an RL algorithm to convergence. RL algorithms are notorious for requiring a large number of environment steps to converge. This is especially crippling when attempting to learn complex behaviors in large, high-dimensional environments. Ho's GAIL paper attempts to remedy precisely this efficiency issue.

Ho's paper is riddled with mathematical intricacies, but at a high-level the key insight they had is this: the process of performing IRL then RL implicitly seeks to produce a policy ($\pi_a$) whose distribution of state-action pairs is \textit{similar} to the state-action pair distribution of the expert policy ($\pi_E$). This process of training a policy to match the state-action distribution of an expert policy can be done using a Generative Adversarial Neural Network (GAN) \citep{GANs}, which is a specific Neural Network architecture built to learn to match an arbitrary distribution. Performing this distribution matching with a GAN instead of with the combination of IRL and RL is significantly more data-efficient because it does not need to run an RL algorithm to convergence during each training iteration. Using a GAN in this way to more-efficiently perform IRL followed by RL is what the authors refer to as 'Generative Adversarial Imitation Learning' (GAIL)\footnote{Specifically, GAIL can be used to learn the same policy that \textit{Maximum Entropy IRL} \citep{ZiebartMaxEntropy} followed by RL would learn.}.

  \begin{figure}[ht]
      \centering
      \includegraphics[width=0.5\textwidth]{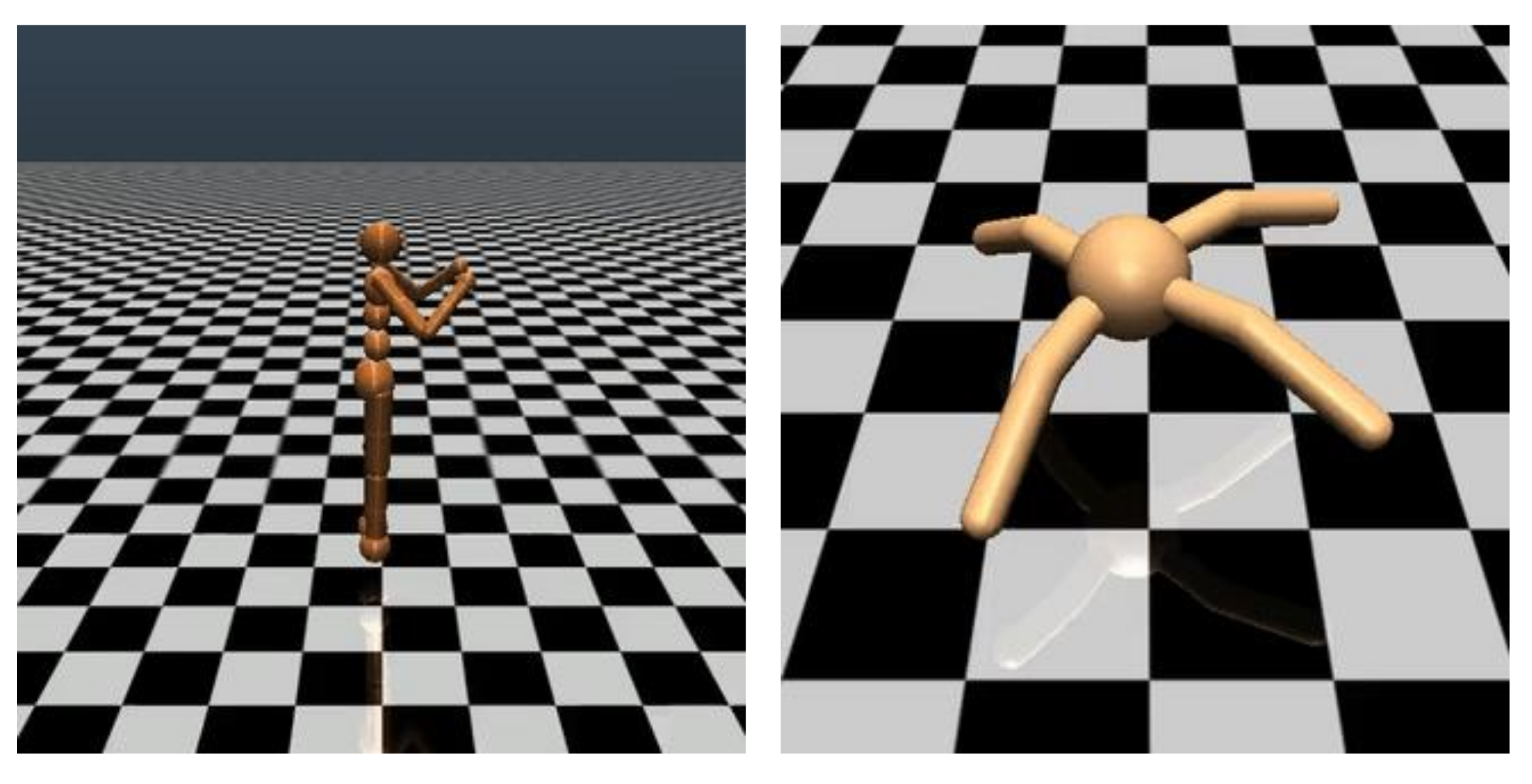}
      \caption{An image from a presentation by one of the authors \protect \footnotemark[6]. GAIL was able to learn to make humanoid and ant walk in simulation with only approximately $50$ timesteps of state-action pairs from a trajectory. The humanoid has approximately $17$ joints and the ant has approximately $8$ joints that the policy must learn to control individually to generate the desired behavior.}
      \label{fig:GAIL}
  \end{figure}
\footnotetext[6]{Link to presentation here: \\ http://efrosgans.eecs.berkeley.edu/CVPR18\_slides/GAIL\_by\_Ermon.pdf}

The paper experimentally demonstrates that GAIL is much more efficient with respect to expert data than 'Behavior Cloning' methods on a number of high-dimensional tasks in simulation. 'Behavior Cloning' methods are nothing but the direct Supervised Learning methods for Imitation Learning used by \citet{Pomerleau:1989:AAL:89851.89891} and discussed in Section \ref{sec:autodrive}. Furthermore, GAIL is compared with two versions of state-of-the-art IRL algorithms, including a version of the algorithm from \citet{Abbeel:2004:ALV:1015330.1015430}. GAIL is shown to achieve superior performance given significantly fewer expert trajectories.

In summary, this paper introduced a novel method called GAIL that can induce the same policy that IRL methods would, but in a more data-efficient manner. Importantly, even though GAIL uses a DNN to learn its policy, it still requires access to the environment just as IRL methods do. This contrasts the Supervised Learning methods (such as Pomerleau's) that can be trained only on the expert demonstrations without needing access to the environment. Furthermore, GAIL is only more efficient than IRL methods in terms of \textit{expert demonstration data}. The authors note that it is not very efficient in terms of the number of interactions required with the environment.

\section{Ghasemipour's Divergence Minimization Perspective Paper}
\label{sec:fmax}
\citet{GAIL} experimentally demonstrated that GAIL is more data-efficient than direct Supervised Learning (or Behavior Cloning). However, they did not provide any strong theoretical justification for why GAIL is able to do this. In fact, as mentioned at the end of Section \ref{sec:IRL}, IRL methods are simply \textit{different} than direct Supervised Learning methods. Aside from some high-level intuitions, it was theoretically unclear why they might offer more robust performance. \citet{ghasemipour2019divergence}'s recent paper sets out to understand the differences between the various approaches to Imitation Learning at a theoretical level to build a unifying perspective amongst them and hopefully use this perspective to develop novel methods.

Section \ref{sec:GAIL} introduced \citet{GAIL}'s insight that IRL methods are really just attempting to learn a policy whose state-action distribution matches the state-action distribution of the expert's policy. Ghasemipour builds on this by proving that \textit{all} Imitation Learning methods are simply attempting to minimize some measure of divergence between the expert's state-action distribution and the learned policy's state-action distribution. Let's use $\rho_{\pi_{E}}$ to denote the distribution of states and actions encountered when following the expert's policy. Similarly, let $\rho_{\pi}$ denote the distribution of states and actions encountered when following the learned policy. Ghasemipour specifically shows that GAIL and related methods seek to minimize the divergence between $\rho_{\pi_{E}}(s_{t},a_{t})$ and $\rho_{\pi}(s_{t},a_{t})$ whereas direct Supervised Learning methods seek to minimize the divergence between $\rho_{\pi_{E}}(a_{t}|s_{t})$ and $\rho_{\pi}(a_{t}|s_{t})$, where $a_{t}$ denotes actions taken at time $t$ and $s_{t}$ denotes the state at time $t$. Ghasemipour experimentally demonstrates that it is precisely this difference - that direct Supervised Learning methods attempt to match the expert's distribution of actions \textbf{conditioned on} states while GAIL and other IRL-based methods attempt to match the expert's \textbf{joint distribution} of actions and states - that makes GAIL perform better than the direct Supervised Learning methods on complex tasks in high-dimensional state spaces.

Additionally, given the insight that \textit{all} Imitation Learning methods just seek to match state-action distributions, Ghasemipour raises the question of whether we should even infer these distributions from expert demonstrations. Instead, he introduces a variant of a state-of-the-art IRL method \citet{AIRL} that he calls FAIRL and then uses a version of this to learn to perform tasks that are directly-specified. Specifically, instead of providing expert demonstrations, Ghasemipour hand-specifies the distribution of states and actions he wants the learned policy to follow.

  \begin{figure}[ht]
      \centering
      \includegraphics[width=0.5\textwidth]{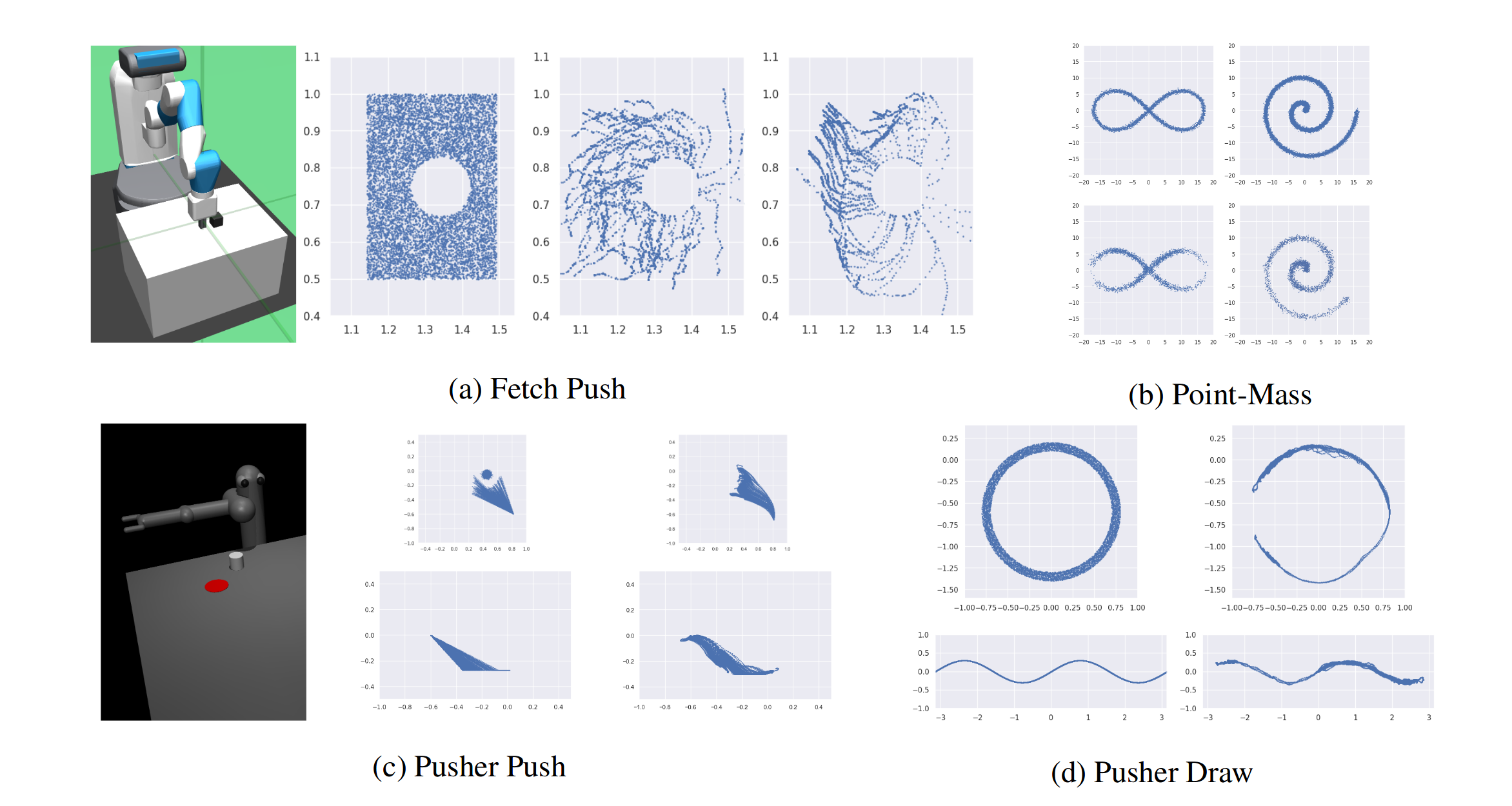}
      \caption{An image from Ghasemipour's paper showing a visualization of experiments performed on learning a policy to match a hand-specified distribution. Subfigures (a) through (d) are visualizations of the distributions for movement actions Ghasemipour specified and corresponding visualizations of the distributinos the learned policies had. The images on the very left showcase the simulation environments for the 'Fetch' (top) and 'Pusher' (bottom) tasks respectively.}
      \label{fig:overview}
  \end{figure}

To summarize, this paper presented a theoretical justification for why recent IRL-based methods (such as GAIL) achieve higher performance on complex Imitation Learning tasks when trained on much fewer expert trajectories than direct Supervised Learning methods. After experimentally verifying their theoretical claim, the authors introduce a new method to perform Imitation Learning without expert trajectories by directly learning to match some hand-specified state-action distribution. In domains where it is easy to hand-specify such distributions (such as simple pushing or movement domains), the author's method is potentially easier to performing Imitation Learning with expert demonstrations. However, it is unlikely that such hand-specification is easy, or even possible, in \textit{all} domains of interest.

\section{Conclusion and Future Outlook}
\label{sec:conclusion}
There has been much progress with Imitation Learning methods over the past $30$ years. Such methods have gone from being able to learn simple, specific tasks given ideal conditions and hand-engineered features (such as \citet{Pomerleau:1989:AAL:89851.89891}'s early success with steering a car) to learning complex tasks in high-dimensional simulation environments (such as block-pushing or teaching an ant to walk) from very few expert demonstrations. Furthermore, we as a community have developed a wide array of methods - ranging from direct Supervised Learning to using Reinforcement Learning - to solve the Imitation Learning problem. These different methods have demonstrated different advantages and disadvantages, and we have recently developed a theoretical understanding of why such differences exist.

Despite all this progress, there are many open questions that remain to be answered and much work that remains to be done. Recent work has shown impressive experimental results in simulation domains, however it remains to be seen whether such methods will transfer well to complex tasks in the real-world. This might be especially difficult because, as noted by \citet{GAIL}, GAIL and related state-of-the-art methods are not that efficient in terms of interaction with the environment needed for them to successfully learn new skills. Furthermore, as \citet{ghasemipour2019divergence} points out, expert demonstrations may not be the easiest way to induce policies for certain tasks. Thus, features of a task that make it easily amenable to solving via Imitation Learning with expert demonstrations should be studied. Additionally, novel ways of providing state-action distributions for a learner to match (for example, via language commands, etc.) can be explored. Finally, while \citet{ghasemipour2019divergence}'s work helped deepen our theoretical understanding of Imitation Learning methods, there is much more to be understood (for example, the exact effect of the measure of divergence chosen (for example, KL versus JS) on Imitation Learning's performance) that could help develop novel methods that can solve increasingly complex tasks in real-world settings.

\pagebreak

\bibliographystyle{plainnat}
\bibliography{references}

\end{document}